\pdfoutput=1
\documentclass[twocolumn,times,final,authoryear]{elsarticle}

\usepackage{prletters}
\usepackage{framed,multirow}

\usepackage{amssymb}
\usepackage{latexsym}

\usepackage{url}
\usepackage{xcolor}
\definecolor{newcolor}{rgb}{.8,.349,.1}

\usepackage{comment}
\usepackage{booktabs}
\usepackage{multirow}
\usepackage{xcolor}
\usepackage[normalem]{ulem}
\usepackage{soul}
\usepackage{subfig}

\newcommand{\ie}{\textit{i.e.}}

\newcommand*{\ra}{\ensuremath{\rightarrow}}

\journal{Pattern Recognition Letters}

\begin{document}
%
% Copyright Notice
\thispagestyle{empty}
\onecolumn
\linespread{1.2}\selectfont{}
{\noindent\Huge Elsevier Copyright Notice}\\[1pt]

{\noindent\large Copyright (c) 2024 Elsevier

\noindent Personal use of this material is permitted. Permission from Elsevier must be obtained for all other uses, in any current or future media, including reprinting/republishing this material for advertising or promotional purposes, creating new collective works, for resale or redistribution to servers or lists, or reuse of any copyrighted component of this work in other works.}\\[1em]

{\noindent\Large Accepted to be published in: Pattern Recognition Letters, 2025.}\\[1in]

{\noindent\large Cite as:}\\[1pt]

{\setlength{\fboxrule}{1pt}
 \fbox{\parbox{0.65\textwidth}{L. F. Alvarenga e Silva, S. F. dos Santos, N. Sebe, and J. Almeida, ``Beyond the Known: Enhancing Open Set Domain Adaptation with Unknown Exploration''. \emph{Pattern Recognition Letters}, 2025, pp. 1--8}}}\\[1in] %, doi: 10.1109/SIBGRAPI54419.2021.00031}}}\\[1in]
 
{\noindent\large BibTeX:}\\[1pt]

{\setlength{\fboxrule}{1pt}
 \fbox{\parbox{0.95\textwidth}{
 @Article\{PRL\_2025\_Silva,
 
 \begin{tabular}{lll}
  & author    & = \{L. F. \{Alvarenga e Silva\} and 
                    S. F. \{dos Santos\} and
                    N. \{Sebe\} and
                    J. \{Almeida\}\},\\
			   
  & title     & = \{Beyond the Known: Enhancing Open Set Domain Adaptation with Unknown Exploration\}, \\
			   
  & journal   & = \{Pattern Recognition Letters\},\\
  
  % & volume    & = \{\},\\
  
  % & number    & = \{\},\\
  
  & pages     & = \{1--8\},\\
  
  % & month     & = \{\},\\
  
  & year      & = \{2025\},\\
  
  % & doi       & = \{10.1109/SIBGRAPI54419.2021.00031\},\\
  \end{tabular}
  
\}
 }}}

\twocolumn
\linespread{1}\selectfont{}
\clearpage

\setcounter{page}{1}

\begin{frontmatter}

\title{Beyond the Known: Enhancing Open Set Domain Adaptation with Unknown Exploration}

\author[1]{Lucas Fernando \surname{Alvarenga e Silva}\corref{cor1}} 
\cortext[cor1]{Corresponding author.
%  Tel.: +0-000-000-0000;  
%  fax: +0-000-000-0000;
}
\ead{lucas.silva@ic.unicamp.br}
\author[2]{Samuel Felipe \surname{dos Santos}}
\ead{samuel.felipe@ufscar.br}
\author[3]{Nicu \surname{Sebe}}
\ead{niculae.sebe@unitn.it}
\author[2]{Jurandy \surname{Almeida}}
\ead{jurandy.almeida@ufscar.br}

%% \affiliation[label1]{organization={},%Department and Organization
%%             citysep={}, % use if no comma needed between city and postcode%%             

\affiliation[1]{organization={Instituto de Computação, Universidade Estadual de Campinas (UNICAMP)},
                addressline={Av. Albert Einstein, 1251}, 
                city={Campinas}, 
                postcode={13083-852}, 
                state={São Paulo},
                country={Brazil}}

\affiliation[2]{organization={Department of Computing, Federal University of São Carlos (UFSCar)},
                addressline={Rod. João Leme dos Santos, km110}, 
                city={São Carlos}, 
                postcode={18052-780}, 
                state={São Paulo},
                country={Brazil}}

\affiliation[3]{organization={Department of Information Engineering and Computer Science, University of Trento (UniTN)},
                addressline={Via Sommarive, 9}, 
                city={Trento},
                postcode={38123}, 
                state={Trentino},
                country={Italy}}

\received{20 April 2024}
\finalform{23 October 2024}
\accepted{13 May 2013}
\availableonline{15 May 2013}
\communicated{S. Sarkar}

\begin{abstract} 
Convolutional neural networks~(CNNs) can learn directly from raw data, resulting in exceptional performance across various research areas. However, factors present in non-controllable environments such as unlabeled datasets with varying levels of domain and category shift can reduce model accuracy. The Open Set Domain Adaptation (OSDA) is a challenging problem that arises when both of these issues occur together. Existing OSDA approaches in literature only align known classes or use supervised training to learn unknown classes as a single new category. In this work, we introduce a new approach to improve OSDA techniques by extracting a set of high-confidence unknown instances and using it as a hard constraint to tighten the classification boundaries. Specifically, we use a new loss constraint that is evaluated in three different ways: (1) using \textit{pristine} negative instances directly; (2) using data augmentation techniques to create randomly \textit{transformed} negatives; and (3) with \textit{generated} synthetic negatives containing adversarial features.
We analyze different strategies to improve the discriminator and the training of the Generative Adversarial Network (GAN) used to generate synthetic negatives.
We conducted extensive experiments and analysis on OVANet using three widely-used public benchmarks, the Office-31, Office-Home, and VisDA datasets. 
We were able to achieve similar H-score to other state-of-the-art methods, while increasing the accuracy on unknown categories.
\end{abstract}

\begin{keyword}
\MSC 41A05\sep 41A10\sep 65D05\sep 65D17
\KWD open set domain adaptation\sep open set recognition\sep domain adaptation

\end{keyword}

\end{frontmatter}

%\linenumbers

\section{Introduction}

Deep Learning (DL) methods have been obtaining astonishing results in several research areas recently, particularly, Convolutional Neural Networks (CNNs) are widely used in computer vision problems~\citep{TPAMI_2020_Geng}.
However, these methods usually operate under unrealistic ideal conditions, where the dataset is fully labeled and belongs to a Closed Set (CS) of categories~\citep{TPAMI_2020_Geng}.
In real-world scenarios, these conditions are not always feasible, and labeling data is often time-consuming, expensive, or even impossible.
As a result, DL methods designed for working with supervised datasets often struggle when faced with unlabeled or partially labeled datasets~\citep{ECCV_2020_Bucci,CVIU_2022_Saltori}.

Two main problems arise in uncontrollable environments: decreased levels of supervision and lack of control over incoming data.
To deal with the decreased level of supervision, a fully annotated dataset (\ie~source domain) similar to the unlabeled target dataset (\ie~target domain) is used to train the model. However, since the data may have different distributions, this may lead to the domain-shift problem~\citep{SIBGRAPI_2021_Silva}.
To deal with the lack of control over the incoming data, the model must be capable of handling examples from possible unknown categories during inference. This is necessary due to the category-shift that may occur between our training and test data as we do not know all data beforehand~\citep{TPAMI_2021_Chen}.
The domain-shift and category-shift problems have their own research areas dedicated to mitigating them individually, the Unsupervised Domain Adaptation~(UDA) and Open Set~(OS) recognition, respectively.
Recently, the more challenging scenario where both domain-shift and category-shift problems occur simultaneously was introduced by \citet{ICCV_2017_Busto}, being named Open Set Domain Adaptation (OSDA).
The OSDA has been widely accepted as a new research field in literature, with many different contributions, including approaches based on adversarial techniques~\citep{ECCV_2018_Saito, ECCV_2020_Rakshit, NC_2020_Gao, WACV_2022_Baktashmotlagh}, Extreme Value Theory (EVT) modeling techniques~\citep{BIGDATA_2022_Xu}, self-supervised techniques~\citep{ECCV_2020_Bucci}, contrastive learning approaches~\citep{Bucci_2022_WACV}, and gradient-based analysis~\citep{ECCV_2022_Liu}.

\begin{figure}[tb]
    \centering
    \includegraphics[width=\linewidth,trim={0 0 0 0},clip]{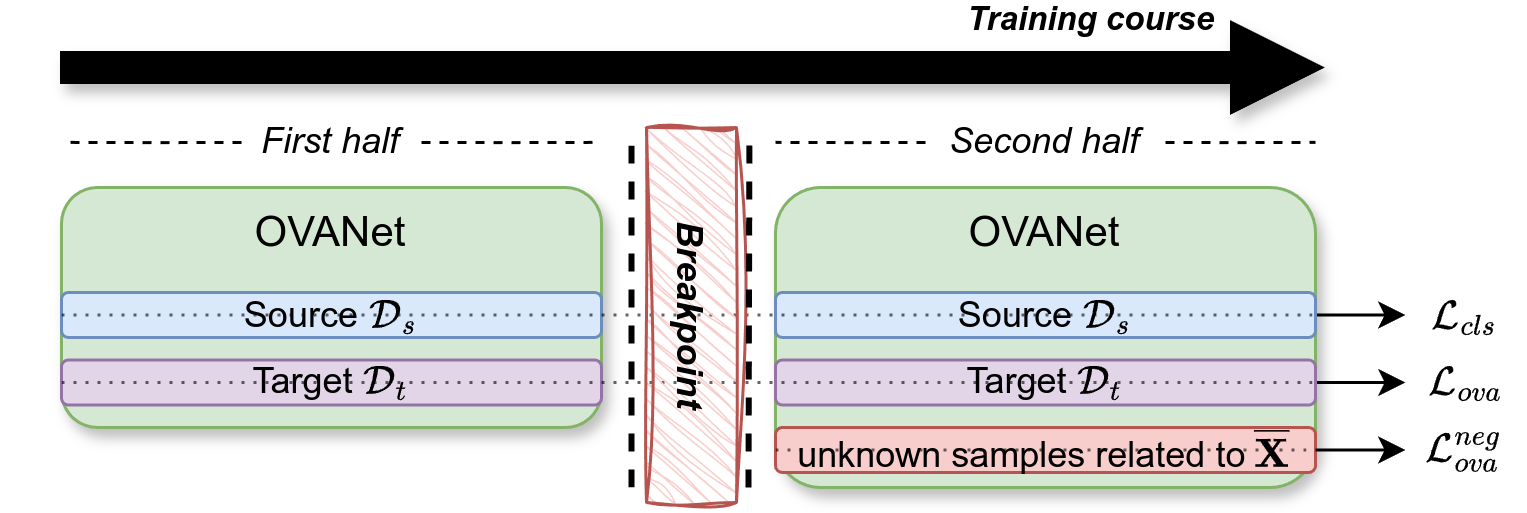}
    \caption{Top-level overview of our approach. Initially, a base method, in this particular case OVANet, is trained for half of the expected iterations. Then, at the \textit{breakpoint}, a sample extraction routine is executed to select the set of the most confident negative samples $\mathbf{\overline{X}}$ from the target domain. Next, the base method leverages the negative samples $\mathbf{\overline{X}}$ as a new constraint $\mathcal{L}_{ova}^{neg}$ for the last half of the iterations to tighten the classification boundaries.}
    \label{fig:overview}
\end{figure}

Usually in OSDA methods, unknown samples in the target domain are rejected, while known samples from the target domain are aligned with the source domain~\citep{LOGHMANI2020198}. 
To achieve this, sets of high-confidence known and unknown samples from the target domain are selected. 
The positive samples from the known set are aligned with the source domain to mitigate the domain shift~\citep{ECCV_2020_Bucci, ECCV_2020_Rakshit, Bucci_2022_WACV}, while the negative samples from the unknown set are usually unexplored during training, being only assigned to an additional logit of the classifier that represents the unknown category and is learned with supervision together with the other classes~\citep{ECCV_2022_Liu}.  
Recent works~\citep{WACV_2022_Baktashmotlagh, ECCV_2022_Liu} leverage negative samples from the unknown set to improve their OSDA approaches, since this data is highly informative, having complicated semantics and possible correlation to known classes that can hinder oversimplified approaches~\citep{ECCV_2022_Liu}.

In light of recent works on closed-set relationships with OS~\citep{ICLR_2022_Vaze} and unknown exploitation~\citep{ECCV_2022_Liu, WACV_2022_Baktashmotlagh}, we hypothesize that the classification performance can be further improved by using unknown samples from the target domain to tighten the boundaries of the closed-set classifier. 
To investigate this hypothesis for the OSDA setting, we propose a three-way extension to OVANet~\citep{Saito_2021_ICCV}, an UNiversal Domain Adaptation~(UNDA) method.
In our method, high-confidence negatives are extracted from the target domain based on a higher confidence threshold~\citep{ECCV_2020_Rakshit} and, then (1) we evaluate the use of a new constraint for the classification of known samples based on the direct use of pure negatives (\textbf{original} approach), (2) we apply data augmentation to these negatives before using them in the aforementioned classification constraint (\textbf{augmentation}, and (3) we train a Generative Adversarial Network~(GAN) model with these negatives to generate negative/adversarial examples, teaching OVANet to reject these synthetic instances that are posed as positives (\textbf{generation} approach). We also conduct ablation studies over the generation approach in order to improve the GAN discriminator and training procedure (\textbf{generation++} approach).
To evaluate our method, we conducted a comprehensive series of experiments on three well-known public datasets for OSDA: Office-31~\citep{ECCV_2010_Saenko}, Office-Home~\citep{CVPR_2017_Venkateswara}, and VisDA~\citep{ARXIV_2017_Peng}.
According to the results we obtained, our methods were able to keep a similar H-score to other state-of-the-art methods in literature, while obtaining significant increases in the recognition of unknown categories.

A preliminary version of this work was presented at the SIBGRAPI - Conference on Graphics, Patterns and Images (SIBGRAPI 2023)~\citep{SIBGRAPI_2023_Silva}.
Here, we introduce several innovations. 
First, a new strategy for generating negative instances by leveraging the already learned knowledge of OVANet in the GAN's discriminator is investigated. Also, a novel training strategy that interleaves the training of OVANet and GAN in the remaining iterations is discussed. In addition, we include new ablation studies, visualizing projections of the features and investigating the sensitivity of our approach to the openness level of the target domain.
Finally, new experiments evaluating the performance of the proposed strategies for the VisDA dataset were added.
The additional strategies and experiments were crucial for the discussions proposed in this paper. 

The remainder of this paper is organized as follows. Section~\ref{sec:related-work} presents related work. Section~\ref{sec:negatives} describes our approach to deal with negative samples. Section~\ref{sec:experiments} presents the experimental setup and results in three datasets. 
Finally, Section~\ref{sec:conclusion} offers conclusions and directions for future research.

\section{Related Work} \label{sec:related-work}

The research areas of OS and UDA can be combined into a more realistic and challenging scenario, known as OSDA, where the distribution of shared categories between the source and target domains is not well-aligned, and some irrelevant categories from the target domain are not present in the source domain.
Although OS~\citep{Neal_2018_ECCV,TPAMI_2021_Chen,ICLR_2022_Vaze,CHAMBERS202275,PRIETO2023113}
and UDA~\citep{WVC_2020_Silva,Saito_2020_NeurIPS,Saito_2021_ICCV,SIBGRAPI_2021_Silva,MARNISSI2022222,CAI2023124} methods are well studied, there are few works for OSDA.

OSDA was first introduced by \citet{ICCV_2017_Busto} and later defined and described by \citet{ECCV_2018_Saito}.
OSDA methods usually have only one source domain~\citep{ECCV_2020_Bucci, ICCV_2017_Busto, ECCV_2018_Saito, BIGDATA_2022_Xu}, but it is possible to extend them to be multi-source~\citep{ECCV_2020_Rakshit, Bucci_2022_WACV}.
A self-supervised model named ROS was introduced by \citet{ECCV_2020_Bucci}. It uses relative-rotation and multi-rotation proxy tasks to narrow the domain gap and learn relevant features from both domains.
\citet{LOGHMANI2020198} proposed a non-negative risk estimator for Positive-Unlabeled (PU) learning and combined it with domain adversarial learning to address the OSDA task.

\citet{Bucci_2022_WACV} also proposed the HyMOS approach to tackle the multi-source OSDA. It uses style transfer to deal with source-to-target alignment and combines contrastive learning and hyperspherical learning to improve the source-to-source alignment.
The MOSDANET is an adversarial method proposed by \citet{ECCV_2020_Rakshit} for the multi-source setting of OSDA.
As noticed by \citet{ECCV_2022_Liu} and \citet{WACV_2022_Baktashmotlagh}, unknown examples have a complex structure but are usually represented by only adding a single new logit to the classifier.
In order to design a feature space that can learn the complex semantics of the unknown samples, \citet{ECCV_2022_Liu} proposed UOL, an approach that makes use of a multi-unknown detector equipped with weight discrepancy constraint and gradient-graph induced annotation.
In a different approach, \citet{WACV_2022_Baktashmotlagh} proposed using a GAN to generate source-like negative samples to be used as a source closed/target supervision.

In general, early OSDA approaches reject unknown examples by using thresholding methods.
While  MOSDANET, ROS, and HyMOS select possibly known examples from the target domain to align them to the source domain and dedicate a single extra logit of the classifier to unknown categories, using unknown examples as a supervision signal to learn it.
In this work, we take a different direction by proposing to focus only on the unknown examples. We extract high-confidence instances from the unknown set of the target domain and use it to tighten the classification boundaries of OVANet.

\section{Dealing with Negatives} \label{sec:negatives}

This section presents our approach. We first describe the confidence measure and extraction procedure used to select a set of probable unknown examples from the target domain. Then, we introduce our strategies, which rely on using original unknown instances, data augmentation, and instance generation. Our methods aim to make use of valuable information from negative samples and improve the class separability of OVANet.

We utilized OVANet~\citep{Saito_2021_ICCV} to evaluate our unknown-oriented learning strategies, which is an UNDA approach that is the foundation of all the research conducted in this work.
OVANet is a recent method that has demonstrated good performance across all four possible class alignments of UNDA, namely: closed-set, partially aligned for closed-set, partially aligned for open-set, and completely open-set fronts. Its source code is freely available on the internet\footnote{\url{https://github.com/VisionLearningGroup/OVANet}}.

In the OSDA problem, we have a source domain $\mathcal{D}_s$ that is labeled and shares a set $L_s$ of categories with a target domain $\mathcal{D}_t$ that is unlabeled. This target domain also includes an additional set $L_{unk}$ of categories not present in $\mathcal{D}_s$, resulting in $L_t = L_s \cup L_{unk}$. 
The source domain can be defined as $\mathcal{D}_s = \{ ( \mathbf{x}_i^{s}, y_i^{s}) \}_{i=1}^{N_s}$, being a set of $N_s$ tuples of a sample $\mathbf{x}_i^{s}$ and its label $y_i^{s}$.
Since the target domain labels are unknown in advance, $\mathcal{D}_t = \{ \mathbf{x}_i^{t} \}_{i=1}^{N_t}$ consists of only $N_t$ target samples. Hence, a sample $\mathbf{x}_i^{t}$ is considered ``known'' if it belongs to any category in $L_s$, otherwise it is ``unknown'' and belongs to $L_{unk}$. The goal of UNDA and OSDA methods is to correctly classify known samples into categories of $L_s$ while rejecting potentially unknown samples that belong to the ``unknown" categories of $L_{unk}$.

\begin{figure}
    \centering
    \includegraphics[width=\linewidth]{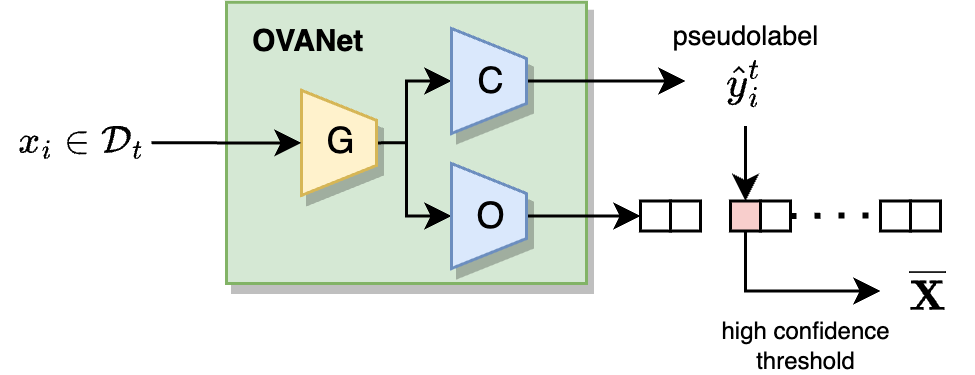}
    \caption{OVANet inference procedure diagram. $G$ refers to the feature extraction entity, in our particular case the ResNet50 model; $C$ consists of the CS head, and $O$ refers to the one-vs-all set of $|L_s|$ binary classifiers of the OS head.}
    \label{fig:inference}
\end{figure}

OVANet was designed to learn tight class boundaries for known samples by exploiting inter-class distance.
As depicted in Figure~\ref{fig:inference}, this method employs a shared feature extractor $G$ for both the closed-set classifier $C$ and the collection of open-set binary classifiers $O$.
The classifier $C$ assigns categories of $L_s$ to the input samples and is trained using cross-entropy loss.
The collection of $|L_s|$ one-vs-all binary classifiers is defined by $O = \{ B_i \}_{i=1}^{|L_s|}$,  where each binary classifier $B_i$ is related to one of the known categories and is trained to recognize samples from that particular category as ``known'' and samples from other categories as ``unknown''. This is achieved thanks to two constraints in the loss function:
\textit{Hard Negative Classifier Sampling} and \textit{Open-set Entropy Minimization} (see Equation~2 of \citet{Saito_2021_ICCV}).
Finally, $p_c(y^k|\mathbf{x}_i)$ is defined as the probability of the classifier $C$ assigning a particular sample $\mathbf{x}_i$ to its $k$-th category. 
Similarly, $p_o(\hat{y}^k|\mathbf{x}_i)$ is defined as the probability of the same sample being considered ``known" by the binary classifier $B_k$ of $O$ and ``unknown" by $B_k$ with probability $1-p_o(\hat{y}^k|\mathbf{x}_i)$.

For inference, an input sample $\mathbf{x}_i^{t}$ is fed to the classifier $C$ that assigns a pseudo-label $\hat{y}_i^{t}$ to it according to the class with maximum posterior probability, \ie, $\hat{y}_i^{t} = \mathrm{argmax}_{k}( p_c(y^k|\mathbf{x}_i^{t}) )$.
Then, the set $O$ of open-set binary classifiers is used to determine whether $\hat{y}_i^{t}$ is ``known'' or ``unknown''.
The pseudo-label $\hat{y}_i^{t}$ is considered ``known'' if $p_o(\hat{y}_i^{t}|\mathbf{x}_i^{t}) \geq 0.5$, otherwise it is ``unknown'' and hence $\mathbf{x}_i^{t}$ is rejected.

Based on the inference procedure of OVANet, we hypothesize that incorporating knowledge from negative samples can enhance OVANet's learning capability and tighten the classification boundaries between known and unknown categories. 
In our approach, we achieve this by training OVANet for only half of the expected iterations, using its knowledge to extract the most confident unknown samples from the target domain, and then utilizing them as an additional negative supervision to fine-tune the binary classifiers in $O$ for the remaining iterations. 
We apply a very-high threshold to define a sample as ``unknown'', \ie, $\mathbf{\overline{X}} = \{ \mathbf{x}_i^{t} | \mathbf{x}_i^{t} \in \mathcal{D}_t, 1-p_o(\hat{y}_i^{t} | \mathbf{x}_i^{t}) > 0.9 \}$, in order to extract a subset $\mathbf{\overline{X}} \subset \mathcal{D}_t$ of likely negative samples.
Finally, we use the negative samples in $\mathbf{\overline{X}}$ as a source of information to compel the binary classifiers in $O$ to tighten the known/unknown classification boundaries. We achieve this by adding a constraint to the loss function of OVANet (see Equation~2 of~\citet{Saito_2021_ICCV}):
\begin{equation} \label{eq:loss_ova_neg}
    \mathcal{L}_{ova}^{neg} = -\frac{1}{|L_s|}\sum_{k=1}^{|L_s|}\log(1-p_o(\hat{y}^k|\mathbf{\overline{x}}_i))
\end{equation}

\begin{figure}[!t]
\centering

\includegraphics[width=\linewidth]{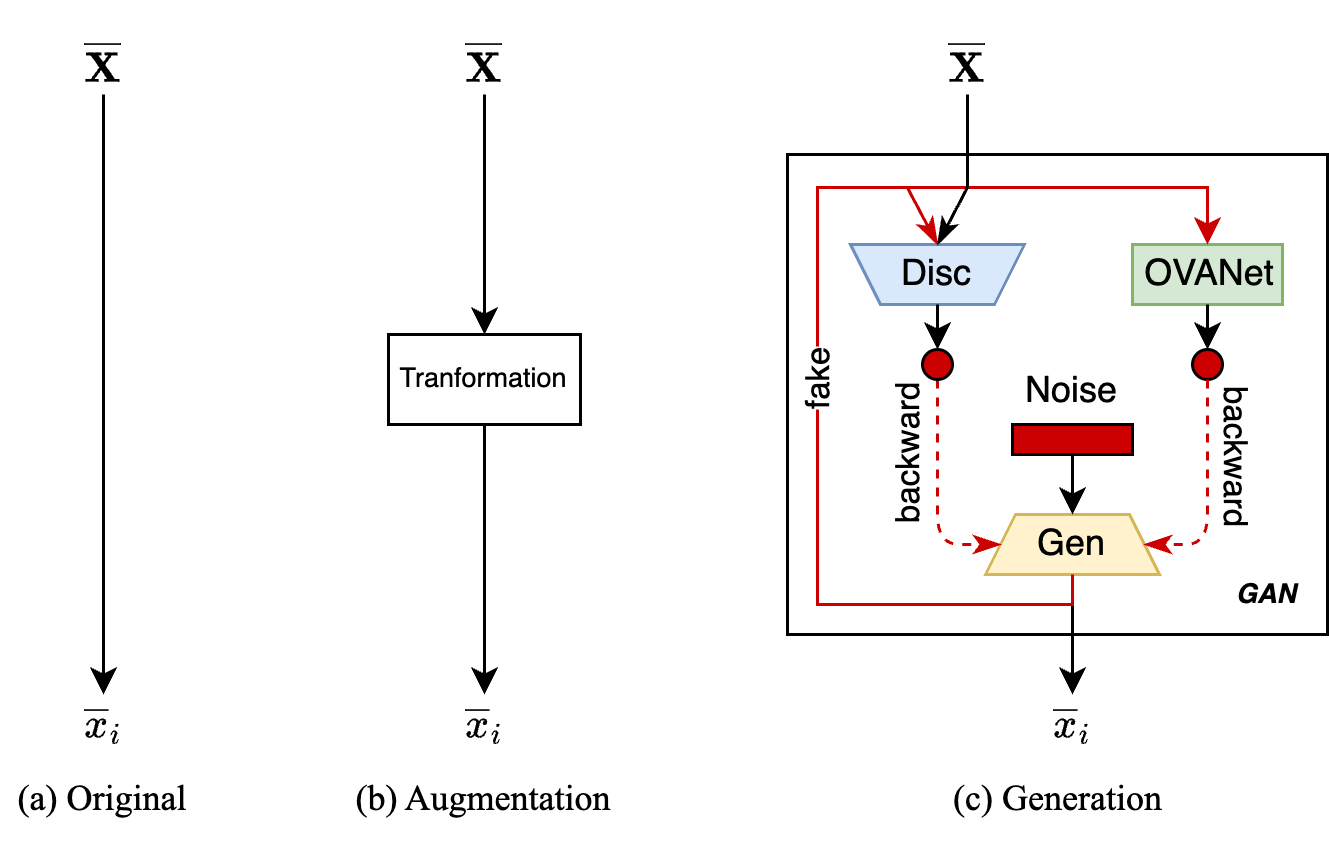}

\caption{Description of all three strategies. (a) refers to the \textit{original} approach, which feeds the instances $\overline{x}_i \in \overline{\mathbf{X}}$ directly to OVANet. (b) depicts the \textit{augmentation} approach, which applies Random Affine and Gaussian Blur transformations to the instances $\overline{x}_i \in \overline{\mathbf{X}}$ before feeding them to OVANet. (c) shows the \textit{generation} approach, where a DCGAN model composed of a generator (Gen) and a discriminator (Disc) is used. Gen is trained to fool both the Disc and OVANet models by outputting synthetic negative instances aligned to the known classes in $L_s$, while Disc distinguishes between real and fake inputs. The dashed arrow indicates the backpropagation step, underscoring the dependence between the Gen entity and the loss error from the Disc and OVANet models.}
\label{fig:approaches}
\end{figure}

We evaluated three different strategies to leverage information from the negative samples in $\mathbf{\overline{X}}$:
\begin{enumerate}
    \item \textbf{Original} (Figure~\ref{fig:approaches}a): Use the negative samples in $\mathbf{\overline{X}}$ to adjust the binary classifiers in $O$.
    \item \textbf{Augmentation} (Figure~\ref{fig:approaches}b): Use data augmentation methods to randomly transform the negative samples in $\mathbf{\overline{X}}$ before feeding them to the binary classifiers in $O$. This maintains the main features of the pristine instances, but introduces small perturbations via the transformations.
    \item \textbf{Generation} (Figure~\ref{fig:approaches}c): Use a GAN to generate synthetic samples that are similar to the negative samples in $\mathbf{\overline{X}}$. To do this, we pause training OVANet after half of the expected iterations, use $\mathbf{\overline{X}}$ to train a GAN for a certain number of epochs, and then continue training OVANet while using the synthetically generated samples to adjust the binary classifiers in $O$. This method allows us to have greater control over features of interest. In general, the GAN consists of two parts, a discriminator and a generator, that are trained jointly and adversarially. The discriminator is usually trained using binary cross-entropy loss to correctly predict whether the input is a real or a fake sample. However, the generator is trained to create samples that are as similar as possible to the real ones to deceive the discriminator's prediction. But, what if we compel the fake samples to deceive the OVANet's actual state, that is, be aligned with a category in $L_s$ by the classifier $C$ and identified as ``known'' by the binary classifiers in $O$? To achieve this, we add two new constraints to the generator's loss function: (\textit{i}) an entropy factor $\mathcal{L}_{ent}^{gen}$ (Equation~\ref{eq:entropy-gen}) based on the classifier $C$, and (\textit{ii}) an agreement factor $\mathcal{L}_{agree}^{gen}$ (Equation~\ref{eq:agreement-gen}) based on the binary classifiers in $O$.
\begin{equation} \label{eq:entropy-gen}
    \mathcal{L}_{ent}^{gen} = -\frac{1}{|L_s|} \sum_{k=1}^{|L_s|} p_c(y^k|\mathbf{\overline{x}}_i)\log(p_c(y^k|\mathbf{\overline{x}}_i))
\end{equation}
\begin{equation} \label{eq:agreement-gen}
    \mathcal{L}_{agree}^{gen} = -\frac{1}{|L_s|}\sum_{k=1}^{|L_s|}\log(p_o(\hat{y}^k|\mathbf{\overline{x}}_i))
\end{equation}
\end{enumerate}

\section{Experiments and Results} \label{sec:experiments}

This section presents our experiments and their results. First, we detail the experimental setup. Then, we discuss ablation studies performed on top of the Generation approach. Finally, we report results on three well-known benchmarks and compare them with state-of-the-art methods.

\subsection{Datasets}

We evaluated our methods and compared them to other state-of-the-art approaches on the Office-31~\citep{ECCV_2010_Saenko}, Office-Home~\citep{CVPR_2017_Venkateswara}, and VisDA~\citep{ARXIV_2017_Peng} datasets.
Office-31 contains 4,652 images from 31 categories of common objects encountered in an office environment. This dataset has a strong imbalance among its Amazon (A), DSLR (D), and Webcam (W) domains. Office-Home presents a similar construct to Office-31. It contains 15,500 images distributed along 4 domains of 65 different categories each. The images of the Art (Ar), Clipart (Cl), Product (Pr), and Real World (Rw) domains were collected from different search engines, presenting a strong domain-shift among them.
The VisDA dataset is focussed on the simulation-to-reality domain shift. It is composed of around 150,000 synthetic images and 50,000 real-world images distributed among 12 categories.

% As suggested by \citet{Saito_2021_ICCV}, the datasets were split into a set of known ($L_s$) and unknown ($L_{unk}$) categories. To do this, the category labels were sorted in alphabetical order, then the first $|L_s|$ categories were chosen as known and the next $|L_{unk}|$ were selected as unknown classes.
Following \citet{Saito_2021_ICCV}, each dataset was split into sets of known ($L_s$) and unknown ($L_{unk}$) categories. To do this, the category labels were sorted in alphabetical order, then the first $|L_s|$ categories were chosen for the known set and the next $|L_{unk}|$ categories were used as the unknown set.

\subsection{Evaluation Metrics}

Previously, the accuracy ($\textnormal{Acc}$) metric, \ie, the proportion of correct classified instances, was the off-the-shelf metric to assess and compare the performance of OSDA methods. However, \citet{ECCV_2020_Bucci} noticed that the strong class imbalance from the target domain of the OS setting may negatively affect the Accuracy metric. Thus, \citet{ECCV_2020_Bucci} propose the H-score ($\textnormal{Hsc}$) metric, a harmonic mean between the accuracy for the known categories ($\textnormal{Acc}^{L_s}$) and the accuracy for the unknown categories ($\textnormal{Acc}^{ L_{unk}}$) given by the Equation~\ref{eq:hscore}. Most of recent works have mainly focused on overall accuracy and H-score. However, one of the primary objectives of OSDA is to reject unknown target samples. For this reason, we choose to report the accuracy for known categories, the accuracy for unknown categories, and the H-score.
\begin{equation} \label{eq:hscore}
    \textnormal{H-score} = 2 \cdot \frac{\textnormal{Acc}^{L_s} \cdot \textnormal{Acc}^{L_{unk}}}{\textnormal{Acc}^{L_s} + \textnormal{Acc}^{L_{unk}}}
\end{equation}

\subsection{Experimental Protocol}

This work follows the same experimental protocol of OVANet~\citep{Saito_2021_ICCV}. For a given dataset, one domain was used as a supervised source domain and another domain as an unsupervised target domain. The training procedure follows a full protocol setting~\citep{ICIAP_2017_Carlucci}, using the known set from the source domain and both the known and unknown sets from the target domain during training, but only the target domain during testing. All experiments were repeated 3 times to ensure statistically sound results, then the mean and standard deviation of the performance metrics were calculated. However, due to space limitations and the relatively low values for the standard deviation (up to 1), we chose to report and discuss only the mean values in Sections~\ref{sec:ablation}~and~\ref{sec:results}.

\subsection{Implementation Details}

The implementation of all our strategies - \ie, Original, Augmentation, and Generation - was built on top of the OVANet's source code.
Initially, the code was updated in order to support recent GPU cards (CUDA 11.7) with a newer PyTorch (1.13.0) version. 
Then, our implementations changed the code as little as possible, ensuring that we ran the same experimental procedure and adopted the authors' hyperparameters and thus performed a fair comparison between OVANet and our strategies. Our code is available at \url{github.com/LucasFernando-aes/UnkE-OVANet}. For evaluation of OVANet in the OSDA setting, \citet{Saito_2021_ICCV} only provided results for the Office-31 dataset (see Table~B of~\citet{Saito_2021_ICCV}). The results for Office-Home were limited to the task Rw\ra Ar (see Table~4 of~\citet{Saito_2021_ICCV}) and the VisDA dataset wasn't evaluated in the OSDA setting. Therefore, we reproduced OVANet results for all datasets using the author's provided scripts to ensure a robust baseline for comparison. In the remainder of this section, we refer to the reproduced results as \textit{Reproducibility}.

A summary of the hyperparameters adopted for each dataset is presented in Table~\ref{tab:hyperparameters}. Common hyperparameters among all datasets are a batch of size 36 and SGD optimizer with a learning rate of $0.01$ for the newly instantiated $C$ and $O$ heads, and $0.001$ for the backbone.

\begin{table}[!htb]
    \centering
    \caption{Hyperparameters adopted for each dataset.}
    \label{tab:hyperparameters}
    \resizebox{\linewidth}{!}{
    \begin{tabular}{r|c|c|c} \toprule
        Hyperparameters                 & Office-31      & Office-Home    %& NaBird          
        & VisDA         \\ \midrule 
        Feature Extractor               & ResNet50       & ResNet50       %& ResNet50        
        & ResNet50      \\
        Source Known Classes            & $|L_s| = 10$     & $|L_s| = 25$     %& $L_s = 300$     
        & $L_s = 6$     \\
        Target Unknown Classes          & $|L_{unk}| = 11$ & $|L_{unk} = 40|$ %& $L_{unk} = 255$ 
        & $L_{unk} = 6$ \\ \midrule
        Training Iterations           & $10000$         & $10000$         %& $10000$          
        & $25000$       \\
        Breakpoint Iterations           & $1000$         & $1000$         %& $1000$          
        & $1000$        \\ \bottomrule
    \end{tabular}
    }
\end{table}

\subsection{Ablation Study}
\label{sec:ablation}

This section describes the ablation studies we performed to develop our approach. First, we analyze the influence of our penalty term $\mathcal{L}_{ova}^{neg}$ for the results of OVANet. Then, we examine how the threshold level affects the selection of unknown instances from the target domain. Next, we tested various strategies to enhance our Generation approach by improving the discriminator and training the GAN in conjunction with OVANet. Finally, we present an analysis of the sensitivity of our approach to the openness of the target domain and a visualization of the features. All experiments in this section were performed on the Office-31 dataset.

\subsubsection{Hyperparameter Tuning}

To control the strength of our penalty $\mathcal{L}_{ova}^{neg}$ (Equation~\ref{eq:loss_ova_neg}), it is weighted by a hyperparameter $\lambda$  before being added to the loss function of OVANet. We evaluated different values for $\lambda$ on the Office-31 dataset, particularly, $\{ 0.01, 0.05, 0.1, 0.2 \}$. The results of these experiments are shown in Table~\ref{tab:lambda}. This analysis has shown that our approach is robust for the evaluated values. Overall, $\lambda = 0.20$ has shown marginally better results and is therefore adopted for the remaining experiments.

\begin{table}[!htb]
    \caption{Analysis of the hyperparameter $\lambda$ that controls the weighting of our penalty $\mathcal{L}_{ova}^{neg}$ to the loss function of OVANet.}    
    \label{tab:lambda}
    \centering
    \resizebox{.8\linewidth}{!}{
    \begin{tabular}{ccccc} \toprule
         \multirow{2}{*}{Approaches}   & \multirow{2}{*}{$\lambda$} & \multicolumn{3}{c}{Average} \\ \cmidrule{3-5}
                                       &                          & Hsc & $\textnormal{Acc}^{L_s}$ & $\textnormal{Acc}^{L_{unk}}$ \\ \midrule
         Reproducibility               & --                       & 91.2        & 90.2                     & 92.6                         \\ \midrule
         \multirow{4}{*}{Original}     & 0.01                     & 91.8        & 90.1                     & 93.7                         \\
                                       & 0.05                     & 91.4        & 89.9                     & 93.2                         \\
                                       & 0.10                     & 91.6        & 89.7                     & 93.8                         \\
                                       & 0.20                     & 92.0        & 90.0                     & 94.4                         \\ \midrule
         \multirow{4}{*}{Augmentation} & 0.01                     & 91.7        & 90.2                     & 93.5                         \\
                                       & 0.05                     & 91.4        & 89.9                     & 93.3                         \\
                                       & 0.10                     & 90.6        & 89.1                     & 92.5                         \\
                                       & 0.20                     & 91.7        & 89.8                     & 94.0                         \\ \midrule
         \multirow{4}{*}{Generation}   & 0.01                     & 91.4        & 91.0                     & 91.9                         \\
                                       & 0.05                     & 91.7        & 90.2                     & 93.5                         \\
                                       & 0.10                     & 91.4        & 90.1                     & 93.1                         \\
                                       & 0.20                     & 91.6        & 90.3                     & 93.2                         \\ \bottomrule
    \end{tabular}
    }
\end{table}

\subsubsection{Threshold Analysis}

\begin{figure}[htb]
\centering

\includegraphics[width=.75\linewidth]{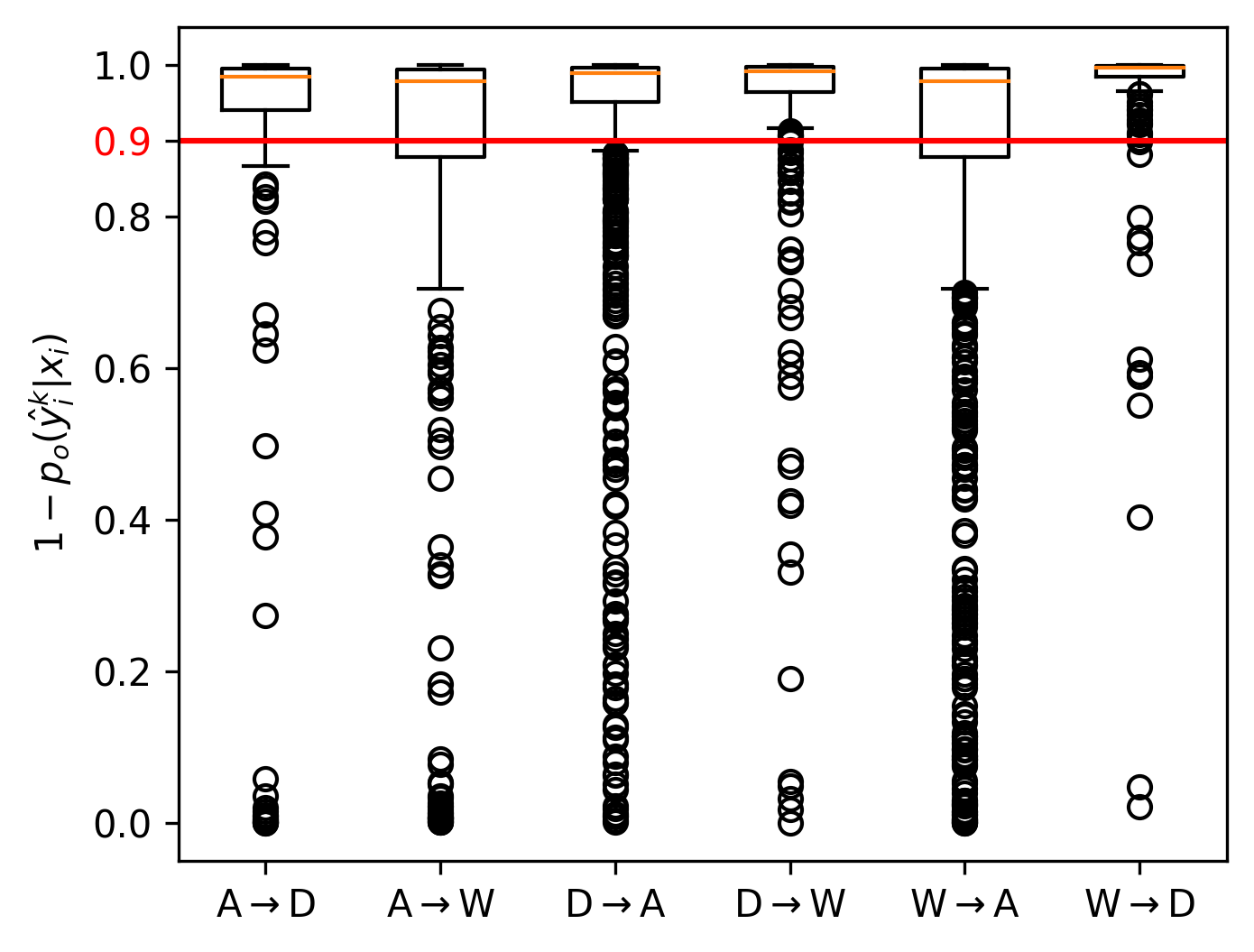}%

\caption{Distribution of the obtained probabilities $1-p_o(\hat{y}_i^{k} | \mathbf{x}_i)$ for unknown instances on the target training set of the Office-31 dataset. Each of the boxplots refers to a specific task and the red line shows the adopted threshold of 0.9.}
\label{fig:approaches:boxplot}
\end{figure}

The primary assumption of this work is that $\mathbf{\overline{X}}$ should contain only unknown instances of the target domain. However, the target domain is not supervised. Thus, in order to construct the set $\mathbf{\overline{X}}$, we proposed to raise the rejection rate of the inference procedure of OVANet (\ie~threshold) to $0.9$, consequentially, only instances with a probability of being unknown greater than $0.9$ will be included in $\mathbf{\overline{X}}$. Figure~\ref{fig:approaches:boxplot} shows that most of the unknown examples are correctly captured, being assigned $1-p_o(\hat{y}_i^{k} | \mathbf{x}_i) \geq 0.9$ by our approach. As we can see, outliers are minimal compared to the vast majority of instances. Therefore, 0.9 is a suitable choice for this threshold.

\subsubsection{Generation Approach Analisys}

In order to further improve our Generation approach, we propose modifications to it. First, we analyze the use of the backbone of the OVANet model as a feature extractor for the discriminator. Then, we investigate the effects of fine-tuning the GAN throughout the final training iterations of OVANet.

\paragraph{GAN's Discriminator Analysis}

Our Generation approach obtained promising results. However, DCGAN is one of the pioneering GANs, containing a very limited and narrow discriminator. Thus, we hypothesize that the GAN can be further enhanced by providing an improved discriminator that leverages the already-learned knowledge from OVANet, \ie, using the feature extractor $G$ from OVANet as the backbone for the discriminator (Figure~\ref{fig:generationpp}). We propose a novel discriminator head to be stacked on top of the shared backbone. Following the DCGAN's discriminator design choice of adopting sigmoid activations, the discriminator's head contains an intermediary linear projection layer (2048\ra 1024 features) and an output linear layer (1024\ra 1 features), both followed by sigmoid activation.

This approach followed the same experimental procedure of the Generation approach (Section~\ref{sec:negatives}), whose results can be observed in Table~\ref{tab:gen_investigation}. 
As we can see, using the feature extractor of OVANet as the backbone of the discriminator yields a slightly lower average Hsc than using the default discriminator from DCGAN. This is due to the small decrease of 2.4\% in the average $\textnormal{Acc}^{L_s}$. However, the average $\textnormal{Acc}^{L_{unk}}$ increased by 2.8\%, compensating the drop in the average $\textnormal{Acc}^{L_s}$.

\begin{table}[!htb]
    \caption{ Classification (\%) results of discriminator analysis for the Generation approach on Office-31. }
    \label{tab:gen_investigation}
    \centering
    \resizebox{\linewidth}{!}{
        \begin{tabular}{ccccccc} \toprule
        \multirow{2}{*}{Tasks}  & \multicolumn{3}{c}{DCGAN's discriminator} & \multicolumn{3}{c}{OVANet's backbone} \\ \cmidrule(lr){2-4} \cmidrule(lr){5-7}
                                & Hsc  & $\textnormal{Acc}^{L_s}$ & $\textnormal{Acc}^{L_{unk}}$ & Hsc  & $\textnormal{Acc}^{L_s}$ & $\textnormal{Acc}^{L_{unk}}$  \\ \midrule
        A\ra D                  & 91.0 & 91.8                     & 90.3                         & 89.6 & 89.0                    & 90.3                          \\
        A\ra W                  & 87.9 & 88.4                     & 87.4                         & 88.9 & 84.9                    & 93.3                          \\
        D\ra A                  & 86.3 & 79.3                     & 94.6                         & 81.9 & 71.2                    & 96.2                          \\
        D\ra W                  & 97.3 & 96.8                     & 97.8                         & 97.4 & 97.1                    & 97.8                          \\
        W\ra A                  & 88.3 & 85.1                     & 91.8                         & 88.8 & 83.9                    & 94.3                          \\
        W\ra D                  & 98.6 & 100.0                    & 97.1                         & 99.8 & 99.6                    & 100.0                         \\ \midrule
        Average                 & \textbf{91.6} & \underline{90.3}& 93.2                         & 91.1 & 87.6                    & \underline{95.3}              \\ \bottomrule
        \end{tabular}
    }
\end{table}

\paragraph{GAN Training Strategy Analysis}

The Generation approach trains the GAN model only once during the training procedure of OVANet, particularly at the breakpoint depicted in Figure~\ref{fig:overview}. At this point, (1) the inference procedure of OVANet is executed and the most probable unknown samples from the target domain are extracted, which (2) are used to train the GAN. After training the GAN, fake examples generated by the GAN are used as negative supervision for training OVANet. However, as the OVANet evolves in the subsequent iterations, the GAN is held frozen with the knowledge acquired from the negative samples in $\mathbf{\overline{X}}$. For this reason, we propose an interleaved training approach to further evolve the GAN as the OVANet naturally evolves in the training course. For this, we assumed the Generation approach with the OVANet's feature extractor as the backbone of the discriminator and repeated steps (1) and (2) after a fixed training interval, jointly evolving the knowledge of the GAN as the OVANet also evolves with the training course.

The interleaved training strategy followed the default experimental procedure of the Generation approach (Section~\ref{sec:negatives}), whose results can be seen in Table~\ref{tab:gen_investigation_training}. It compares the results of training the GAN only once at the breakpoint and the proposed interleaved training. As depicted, the Interleaved training obtained a slightly better Hsc than the single training, having an accuracy 2\% higher in known categories and 1.2\% lower in unknown categories. Given the good compromise between the $\textnormal{Acc}^{L_{unk}}$ and $\textnormal{Acc}^{L_s}$ metrics, we provide results of this Generation approach improvement as \textit{Generation++} in next sections.

\begin{table}[!htb]
    \caption{ Classification (\%) results of training strategy for the Generation approach on Office-31. }
    \label{tab:gen_investigation_training}
    \centering
    \resizebox{\linewidth}{!}{
        \begin{tabular}{ccccccc} \toprule
        \multirow{2}{*}{Tasks}  & \multicolumn{3}{c}{Single training} & \multicolumn{3}{c}{Interleaved training} \\ \cmidrule(lr){2-4} \cmidrule(lr){5-7}
                                & Hsc  & $\textnormal{Acc}^{L_s}$ & $\textnormal{Acc}^{L_{unk}}$ & Hsc  & $\textnormal{Acc}^{L_s}$ & $\textnormal{Acc}^{L_{unk}}$  \\ \midrule
        A\ra D                  & 89.6 & 89.0                    & 90.3                          & 90.5 & 91.2                     & 89.7                          \\
        A\ra W                  & 88.9 & 84.9                    & 93.3                          & 88.7 & 88.2                     & 89.2                          \\
        D\ra A                  & 81.9 & 71.2                    & 96.2                          & 84.4 & 75.5                     & 95.7                          \\
        D\ra W                  & 97.4 & 97.1                    & 97.8                          & 97.5 & 97.2                     & 97.8                          \\
        W\ra A                  & 88.8 & 83.9                    & 94.3                          & 89.8 & 85.9                     & 94.1                          \\
        W\ra D                  & 99.8 & 99.6                    & 100.0                         & 98.9 & 99.6                     & 98.3                          \\ \midrule
        Average                 & 91.1 & 87.6                    & \underline{95.3}                          & \textbf{91.6} & \underline{89.6}& 94.1                          \\ \bottomrule
        \end{tabular}
    }
\end{table}

\subsubsection{Openness Analysis}

\begin{figure}[htb]
    \begin{minipage}[t]{0.48\linewidth}
        \centering
        \includegraphics[scale=.11]{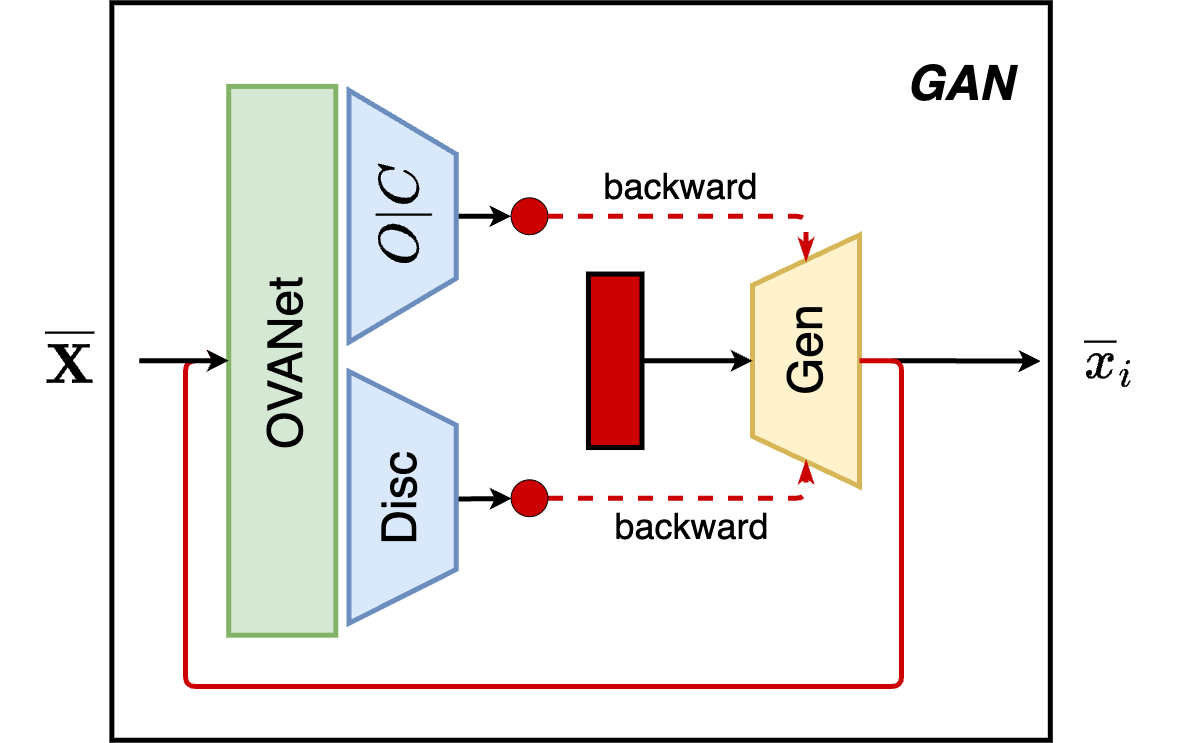}
        \caption{Ilustrations of the OVANet's feature extraction as the backbone of the GAN discriminator.}
        \label{fig:generationpp}
    \end{minipage}%
    \hfill
    \begin{minipage}[t]{0.48\linewidth}
        \centering
        \includegraphics[scale=.3]{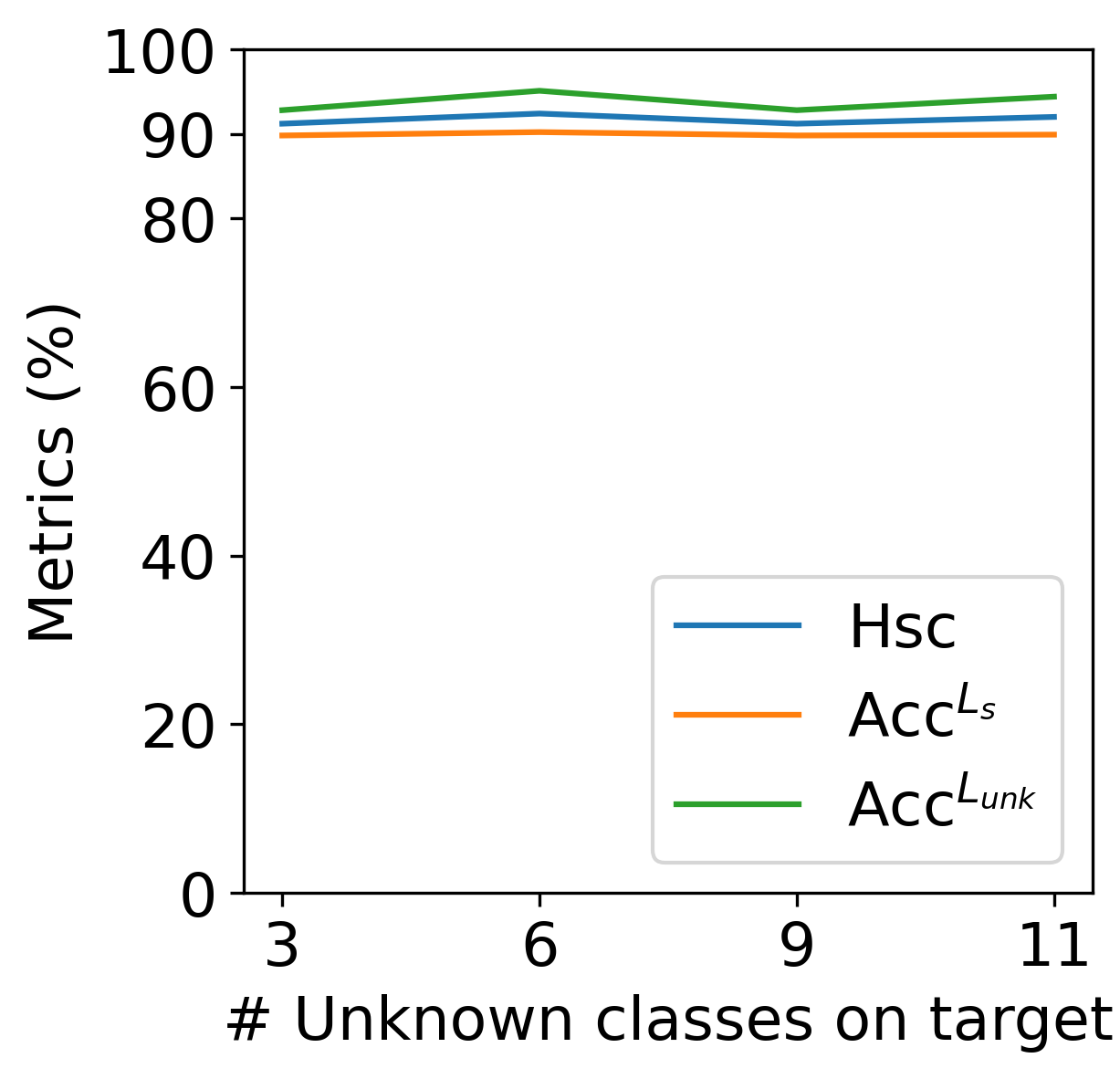}
        \caption{Openness analysis of the Generation++ approach.}
        \label{fig:openness}
    \end{minipage}
\end{figure}

We also evaluated the Generation++ approach sensitivity regarding the openness level of the target domain. For this, we varied the number of unknown categories from the target domain in $L_{unk}$ by $\{ 3, 6, 9, \textnormal{and}, 11 \}$ for the Office-31 dataset and measured the performance metrics obtained for the Generation++ approach. Figure~\ref{fig:openness} shows the Hsc, $\textnormal{Acc}^{L_s}$, and $\textnormal{Acc}^{L_{unk}}$ results averaged across all tasks for each of the openness level. Particularly, the lines are roughly stable, showing that our approach is robust to the openness level of the target domain.

\subsection{Results and Discussion} \label{sec:results}

This section presents the results for the \textit{Original}, \textit{Augmentation}, \textit{Generation}, and \textit{Generation++} approaches. First, we compare the classification performance among these strategies. Then, we position our approach in relation to state-of-the-art methods on the Office-31, Office-Home, and VisDA datasets.

Table~\ref{tab:visda-and-officehome} presents the average Hsc, $\textnormal{Acc}^{L_s}$ and $\textnormal{Acc}^{L_{unk}}$ results for our strategies on the Office-Home and VisDA datasets. The results for Office-31 were discussed in Section~\ref{sec:ablation}.
In general, the Original, Augmentation, and Generation approaches achieved Hsc results better than the Reproducibility baseline, with Original achieving the best performance with 0.8\% of improvements on average in Office-Home. However, our approach is able to provide a higher level of $\textnormal{Acc}^{L_{unk}}$ in exchange for small amounts of $\textnormal{Acc}^{L_s}$ percentage points. For example, the Original and Augmentation approaches yield improvements of 4.5 and 2.7 percentage points of $\textnormal{Acc}^{L_{unk}}$ in exchange for, respectively, 1.7 and 1.3 percentage points of $\textnormal{Acc}^{L_s}$.
The same pattern is observed for the VisDA dataset. The Generation++ approach yields marginally better Hsc results, but when focusing on $\textnormal{Acc}^{L_{unk}}$, the Generation approach presented 4.9 percentage points of improvements in exchange for 3.6 of $\textnormal{Acc}^{L_s}$.

Table~\ref{tab:results} compares the average Hsc obtained by our approaches and other methods from the literature for the Office-31, Office-Home, and VisDA datasets. Particularly, literature results were reported by \citet{AAAI_2024_Lu}, which used the same experimental protocol of OVANet, ensuring a fair comparison with the results of our approaches.
%Results highlighted in bold refer to the best-case results of our approaches for the respective task; underscored results were the best results among all methods.
%
Roughly speaking, the results depend on the task at hand, with no one-size-fits-all approach for the OSDA problem.
Among our strategies (the bottom part of the table), exploiting unknown examples during training has shown promising results, with numerous best-result cases attained to one of our strategies. Particularly, Original provided the best average Hsc for Office-31 (92.0\%); Augmentation provided the best performance for Office-Home (64.5\%), and Generation++ provided the best performance for VisDA (61.0\%).
However, even though our approaches provided results better than the Reproducibility baseline, they only refer to the best on the tasks W\ra A and W\ra D, achieving 89.8\% for Generation++ and 99.4 for Original and Augmentation approaches, respectively.
Yet, the unknown exploitation could improve OVANet, a 2021 approach, to be barely comparable with novel state-of-the-art methods, such as MLNet~\citep{AAAI_2024_Lu}, a 2024 approach, and NCAL~\citep{PR_2023_Su}, a 2023 approach.

Particularly, our strategies are method-agnostic, meaning it may be possible to adapt them to any underlying method, further improving its results.

\begin{table}[htb]
    \caption{Classification (\%) results for the Office-Home and VisDA datasets.}
    \label{tab:visda-and-officehome}
    \centering
    \resizebox{\linewidth}{!}{
    \begin{tabular}{ccccccc}  \toprule
        \multirow{2}{*}{Approaches} & \multicolumn{3}{c}{Office-Home}                                & \multicolumn{3}{c}{VisDA}                                       \\ \cmidrule(lr){2-4} \cmidrule(lr){5-7}
                                    & Hsc  & $\textnormal{Acc}^{L_s}$ & $\textnormal{Acc}^{L_{unk}}$ & Hsc  & $\textnormal{Acc}^{L_s}$ & $\textnormal{Acc}^{L_{unk}}$ \\ \midrule
        Reproducibility             & 63.7          & \underline{61.8}         & 69.4                         & 60.9          & 52.0                     & 73.3                         \\ \midrule
        Original                    & \textbf{64.5} & 60.1                     & 73.2                         & 59.4          & 50.5                     & 72.2                         \\
        Augmentation                & 64.2          & 60.5                     & 72.1                         & 60.1          & 50.7                     & 74.7                         \\
        Generation                  & 63.7          & 61.2                     & 70.6                         & 59.7          & 48.4                     & \underline{78.2}             \\ 
        Generation++                & 63.0          & 55.9                     & \underline{76.4}             & \textbf{61.0} & \underline{53.7}         & 70.7                         \\ \bottomrule
    \end{tabular}%
    }
\end{table}

\begin{table*}[!htb]
    \caption{ Summary of H-Score (\%) results for the Office-31, Office-Home and VisDA datasets.}
    \label{tab:results}
    \centering
    \resizebox{\textwidth}{!}{%
        \begin{tabular}{cccccccccccccccccccccc} \toprule
        \multirow{2}{*}{Approaches} & \multicolumn{7}{c}{Office-31} & \multicolumn{13}{c}{Office-Home} & \multirow{2}{*}{VisDA} \\ \cmidrule(lr){2-8} \cmidrule(lr){9-21}  
                        & A\ra D & A\ra W & D\ra A & D\ra W & W\ra A & W\ra D & Avg   & Ar\ra Cl & Ar\ra Pr & Ar\ra Re & Cl\ra Ar & Cl\ra Pr & Cl\ra Re & Pr\ra Ar & Pr\ra Cl & Pr\ra Re & Re\ra Ar & Re\ra Cl & Re\ra Pr & Avg  & \\ \cmidrule(lr){1-1} \cmidrule(lr){2-8} \cmidrule(lr){9-21} \cmidrule(lr){22-22}
        UAN             & 38.9   & 46.8   & 68.0   & 68.8   & 54.9   & 53.0   & 55.1  & 40.3     & 41.5     & 46.1     & 53.2     & 48.0     & 53.7     & 40.6     & 39.8     & 52.5     & 53.6     & 43.7     & 56.9     & 47.5 & 51.9  \\ 
        CMU             & 52.6   & 55.7   & 76.5   & 75.9   & 65.8   & 64.7   & 65.2  & 45.1     & 48.3     & 51.7     & 58.9     & 55.4     & 61.2     & 46.5     & 43.8     & 58.0     & 58.6     & 50.1     & 61.8     & 53.3 & 54.2  \\ 
        DANCE           & 84.9   & 78.8   & 79.1   & 78.8   & 68.3   & 88.9   & 79.8  & 61.9     & 61.3     & 63.7     & 64.2     & 58.6     & 62.6     & \underline{67.4}     & 61.0     & 65.5     & 65.9     & 61.3     & 64.2     & 63.1 & 67.5  \\ 
        DCC             & 58.3   & 54.8   & 67.2   & 89.4   & 85.3   & 80.9   & 72.7  & 56.1     & 67.5     & 66.7     & 49.6     & 66.5     & 64.0     & 55.8     & 53.0     & 70.5     & 61.6     & 57.2     & 71.9     & 61.7 & 59.6  \\ 
        GATE            & 88.4   & 86.5   & 84.2   & 95.0   & 86.1   & 96.7   & 89.5  & 63.8     & 70.5     & 75.8     & 66.4     & 67.9     & 71.7     & 67.3     & \underline{61.5}     & 76.0     & 70.4     & 61.8     & 75.1     & 69.0 & 70.8  \\ 
        TNT             & 85.8   & 82.3   & 80.7   & 91.2   & 81.5   & 96.2   & 86.3  & 63.4     & 67.9     & 74.9     & 65.7     & 67.1     & 68.3     & 64.5     & 58.1     & 73.2     & 67.8     & 61.9     & 74.5     & 67.3 & \underline{71.6}  \\ 
        NCAL            & 84.0   & \underline{93.4}   & \underline{93.4}   & 85.4   & 89.0   & 87.2   & 88.7  & \underline{64.2}     & \underline{74.1}     & \underline{80.5}     & \underline{68.1}     & \underline{72.5}     & \underline{77.0}     & 66.9     & 58.1     & \underline{79.1}     & \underline{74.6}     & \underline{63.5}     & \underline{79.6}     & \underline{71.5} & 69.3  \\ 
        MLNet           & \underline{92.5}   & 91.4   & 87.3   & \underline{98.0}   & 87.5   & 98.9   & \underline{92.6}  & 61.3     & 69.9     & 74.4     & 63.1     & 68.2     & 70.4     & 62.0     & 59.9     & 72.4     & 69.1     & 62.6     & 71.1     & 67.0 & 63.9  \\ \cmidrule(lr){1-1} \cmidrule(lr){2-8} \cmidrule(lr){9-21} \cmidrule(lr){22-22}
        Reproducibility & 90.3   & 87.4   & 84.8   & \textbf{97.8}   & 87.9   & 98.8   & 91.2  & \textbf{58.5}     & 66.2     & 69.8     & 60.9     & 64.6     & 67.9     & \textbf{59.}9     & \textbf{53.6}     & 69.4     & 68.3     & 58.6     & 66.6     & 63.7 & 60.9  \\ 
        Original        & \textbf{91.0}   & 87.6   & \textbf{87.6}   & 97.3   & 89.1   & \textbf{\underline{99.4}}   & \textbf{92.0}  & 58.3     & \textbf{68.5}     & 72.3     & 61.3     & \textbf{66.0}     & \textbf{69.5}     & 58.9     & 52.0     & 70.0     & 68.2     & \textbf{58.8}     & \textbf{69.6}     & \textbf{64.5} & 59.4  \\
        Augmentation    & 90.3   & 88.0   & 86.0   & 97.7   & 88.5   & \textbf{\underline{99.4}}   & 91.7  & 57.7     & 67.0     & 71.7     & \textbf{61.6}     & 65.8     & 68.4     & 59.2     & 52.2     & 70.4     & \textbf{68.5}     & 58.7     & 68.7     & 64.2 & 60.1  \\
        Generation      & \textbf{91.0}   & 87.9   & 86.3   & 97.3   & 88.3   & 98.6   & 91.6  & 58.1     & 66.6     & 69.4     & 60.6     & 65.1     & 68.7     & 59.8     & 52.4     & 69.7     & 68.0     & 58.8     & 67.5     & 63.7 & 59.7  \\
        Generation++    & 90.5   & \textbf{88.7}   & 84.4   & 97.5   & \textbf{\underline{89.8}}   & 98.9   & 91.6  & 55.5     & 66.9     & \textbf{73.2}     & 59.1     & 65.4     & 69.2     & 57.2     & 46.4     & \textbf{70.5}     & 67.5     & 54.3     & 71.4     & 63.0 & \textbf{61.0} \\ \bottomrule 
        \end{tabular}
    }
\end{table*}

\subsubsection{Visualization Analysis}

\begin{figure}[htb]
    \begin{minipage}[t]{0.26\linewidth}
        \centering
        \includegraphics[width=\linewidth]{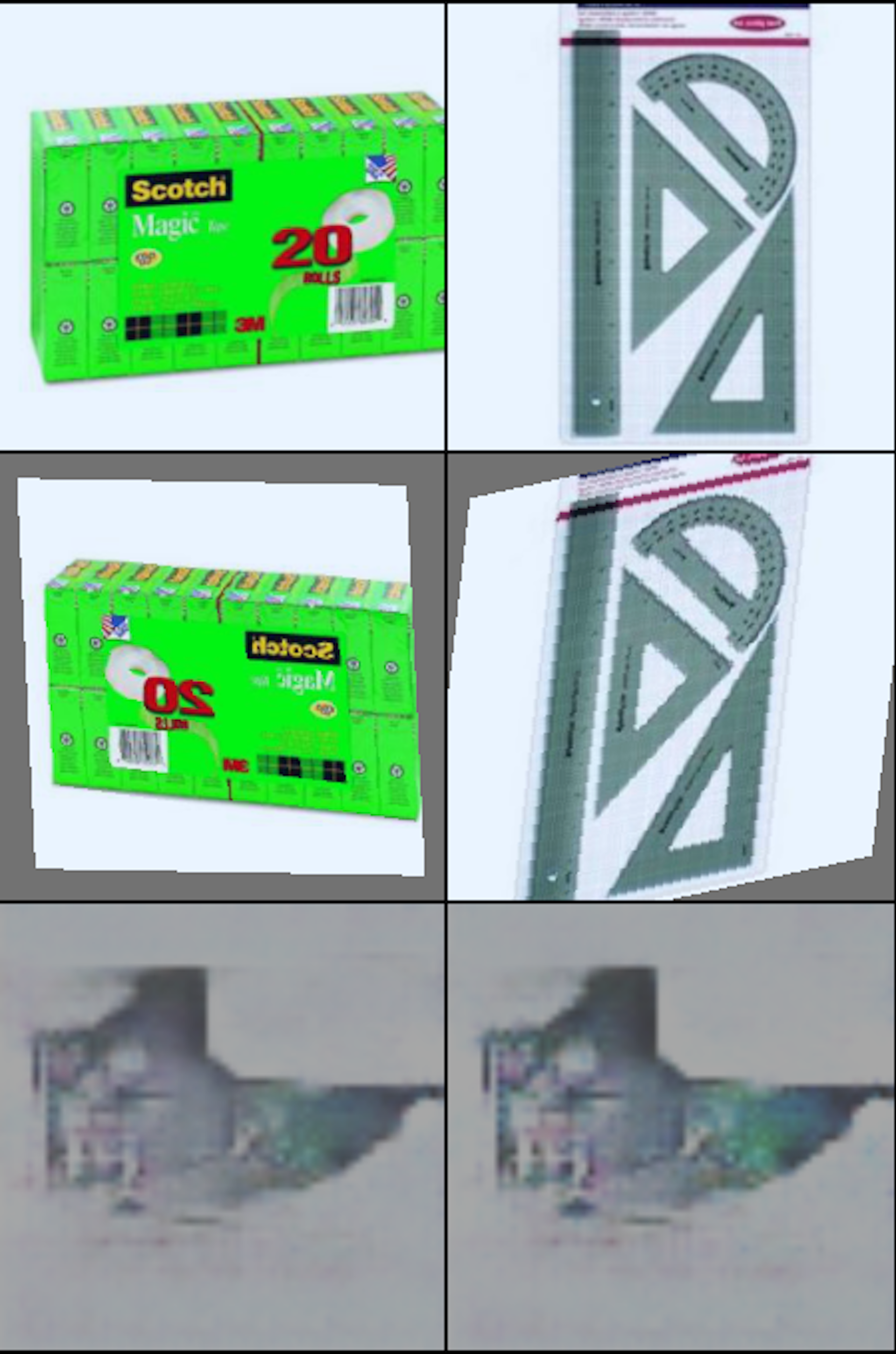}
        
        (a)

    \end{minipage}%
    \hfill
    \begin{minipage}[t]{0.74\linewidth}
        \centering
        \includegraphics[width=.9\linewidth]{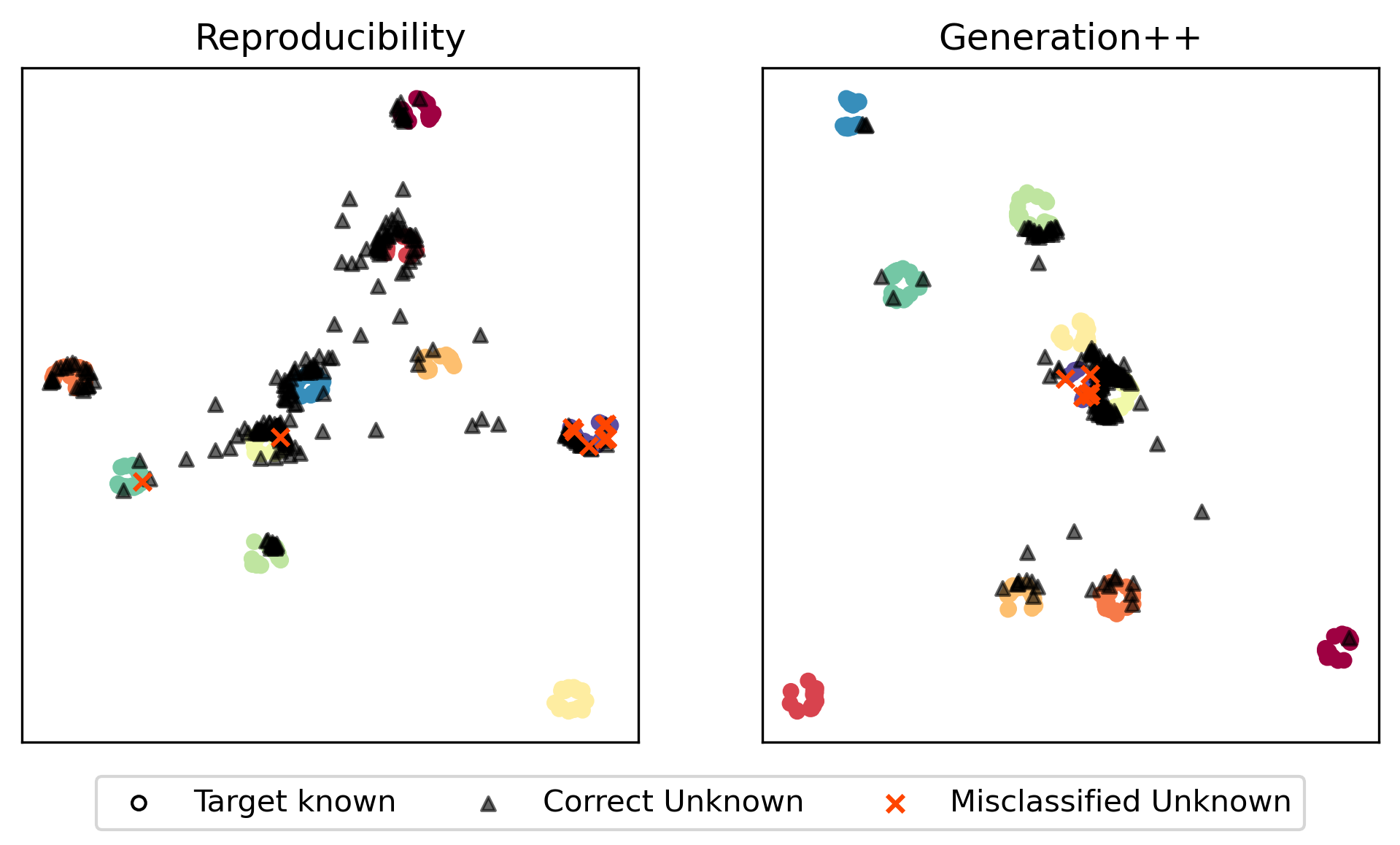}
        
        (b)

    \label{fig:umap}
    \end{minipage}
    \caption{Visual inspection of the proposed method. (a) displays two examples per row for the  Amazon domain of Office-31. From top to bottom, each row shows the original, augmented, and generated instance. (b) shows the UMAP projection of a 2048-dimensional feature space into a 2-dimensional space for the task D\ra W of Office-31. The left-hand plot depicts the projections for Reproducibility, while the right-hand plot shows those for Generation++.}
    \label{fig:visualizations}
\end{figure}

In Figure~\ref{fig:visualizations}, we show some visualization aspects of our approach. Figure~\ref{fig:visualizations}~(a) contains negative instances for the Original, Augmentation, and Generation approaches. The synthetically generated image of the third row is not meaningful at all, which is expected since Equations~\ref{eq:entropy-gen}~and~\ref{eq:agreement-gen} are added to the loss function of DCGAN.
Figure~\ref{fig:visualizations}~(b) shows the UMAP~\citep{CORR_2018_McInnes} projections of the features extracted by the ResNet50 backbone for the Reproducibility and Generation++ experiments\footnote{Due to space limitations, only the projection for the task D\ra W was shown.}. In particular, the projections of Reproducibility already present a certain level of unknown samples being projected to class boundaries. However, it also contains a considerable amount of unknown samples sparsely distributed in the projection space, making it difficult to delimit the classification boundaries. On the other hand, Generation++ clearly shows the motivation behind Equations~\ref{eq:entropy-gen}~and~\ref{eq:agreement-gen}, respectively, by reducing the sparsely distributed unknown instances by pushing their projections to the boundary of some known class, thus tightening the classification boundaries.

\section{Conclusion} \label{sec:conclusion}

Normally, DL models struggle when deployed to real-world applications, suffering from the use of unsupervised datasets, DA, and OS recognition problems. For this reason, the OSDA research area aims to develop DL models that solve these problems, identifying and grouping unknown examples from the target domain that possibly appear during inference. In this work, we hypothesize that tightening classification boundaries of OSDA methods may help the identification of unknown examples, thus, we propose the Original, Augmentation, and Generation approaches to improve OSDA methods. The use of our strategies may achieve similar Hsc to other state-of-the-art methods, while increasing the $\textnormal{Acc}^{L_{unk}}$ in exchange for a small amount of $\textnormal{Acc}^{L_{unk}}$. Particularly, the Original approach improves $\textnormal{Acc}^{L_{unk}}$ by 4.5\% in exchange for 1.7\% of $\textnormal{Acc}^{L_s}$ on Office-Home. Moreover, in Section~\ref{sec:ablation} we identified and evaluated optimization opportunities in the GAN module and the training procedure of the Generation approach; and evaluated the OSDA resulting approaches in a new task of OS recognition with VisDA dataset, further advancing the performance of the OSDA model. In future work, we expect to investigate the use of recent image generation techniques such as diffusion models and different OSDA methods.

\section*{Acknowledgments}
This work was supported by São Paulo Research Foundation - FAPESP (grants 2013/08293-7, 2019/17874-0, 2020/08770-3, 2021/13348-1, 2023/03328-9, 2023/17577-0), FAPESP-Microsoft Research Institute (grant 2017/25908-6), National Council for Scientific and Technological Development - CNPq (grants 315220/2023-6 and 420442/2023-5), and LNCC via resources of the SDumont supercomputer of the IDeepS project.

% \bibliographystyle{model5-names}
% \bibliography{refs}

\end{document}